\documentclass[sigconf]{acmart}

\copyrightyear{2025}
\acmYear{2025}
\setcopyright{acmlicensed}\acmConference[MM '25]{Proceedings of the 33rd ACM International Conference on Multimedia}{October 27--31, 2025}{Dublin, Ireland}
\acmBooktitle{Proceedings of the 33rd ACM International Conference on Multimedia (MM '25), October 27--31, 2025, Dublin, Ireland}
\acmDOI{10.1145/3746027.3754958}
\acmISBN{979-8-4007-2035-2/2025/10}

\AtBeginDocument{
  }

\usepackage{makecell}
\usepackage{multirow}
\usepackage{colortbl}
\usepackage{color}
\usepackage{hyperref}
\usepackage{amsmath}
\usepackage{cleveref}
\usepackage[english]{babel}

\newcommand{\bc}{\mathbf{c}}

\newcommand{\bI}{\mathbf{I}}

\newcommand{\bp}{\mathbf{p}}

\newcommand{\bx}{\mathbf{x}}

\newcommand{\RRR}{\mathbb{R}}

\newcommand{\bmu}{\boldsymbol{\mu}}

\definecolor{susungcolor}{RGB}{0, 50, 150}

\hypersetup{
    colorlinks=true,    
}
\definecolor{c1}{HTML}{ff4777}

\begin{document}

\title{Re-Activating Frozen Primitives for 3D Gaussian Splatting} 

\author{Yuxin Cheng}
\email{yxcheng@connect.hku.hk}
\orcid{0000-0002-3826-9844}
\affiliation{
  \institution{The University of Hong Kong}
  \city{Hong Kong SAR}
  \country{China}
}
\author{Binxiao Huang}
\email{huangbx7@connect.hku.hk}
\orcid{0000-0001-5316-703X}
\affiliation{
  \institution{The University of Hong Kong}
  \city{Hong Kong SAR}
  \country{China}
}
\author{Wenyong Zhou}
\email{wenyongz@connect.hku.hk}
\orcid{0009-0008-0427-7935}
\affiliation{
  \institution{The University of Hong Kong}
  \city{Hong Kong SAR}
  \country{China}
}
\author{Taiqiang Wu}
\email{takiwu@connect.hku.hk}
\orcid{0000-0002-3664-3513}
\affiliation{
  \institution{The University of Hong Kong}
  \city{Hong Kong SAR}
  \country{China}
}
\author{Zhengwu Liu}
\email{zwliu@eee.hku.hk}
\orcid{0000-0001-7968-9469}
\affiliation{
  \institution{The University of Hong Kong}
  \city{Hong Kong SAR}
  \country{China}
}
\author{Graziano Chesi}
\email{chesi@eee.hku.hk}
\orcid{0000-0003-4214-4224}
\affiliation{
  \institution{The University of Hong Kong}
  \city{Hong Kong SAR}
  \country{China}
}
\author{Ngai Wong}
\authornote{Corresponding author.}
\email{nwong@eee.hku.hk}
\orcid{0000-0002-3026-0108}
\affiliation{
  \institution{The University of Hong Kong}
  \city{Hong Kong SAR}
  \country{China}
}

\renewcommand{\shortauthors}{Yuxin Cheng et al.}

\begin{abstract}
3D Gaussian Splatting (3D-GS) achieves real-time photorealistic novel view synthesis, yet struggles with complex scenes due to over-reconstruction artifacts, manifesting as local blurring and needle-shape distortions. While recent approaches attribute these issues to insufficient splitting of large-scale Gaussians, we identify two fundamental limitations: gradient magnitude dilution during densification and the primitive frozen phenomenon, where essential Gaussian densification is inhibited in complex regions while suboptimally scaled Gaussians become trapped in local optima. To address these challenges, we introduce ReAct-GS, a method founded on the principle of re-activation. Our approach features: (1) an importance-aware densification criterion incorporating $\alpha$-blending weights from multiple viewpoints to re-activate stalled primitive growth in complex regions, and (2) a re-activation mechanism that revitalizes frozen primitives through adaptive parameter perturbations. Comprehensive experiments across diverse real-world datasets demonstrate that ReAct-GS effectively eliminates over-reconstruction artifacts and achieves state-of-the-art performance on standard novel view synthesis metrics while preserving intricate geometric details. Additionally, our re-activation mechanism yields consistent improvements when integrated with other 3D-GS variants such as Pixel-GS, demonstrating its broad applicability.
\end{abstract}



\begin{teaserfigure}
  \includegraphics[width=\textwidth]{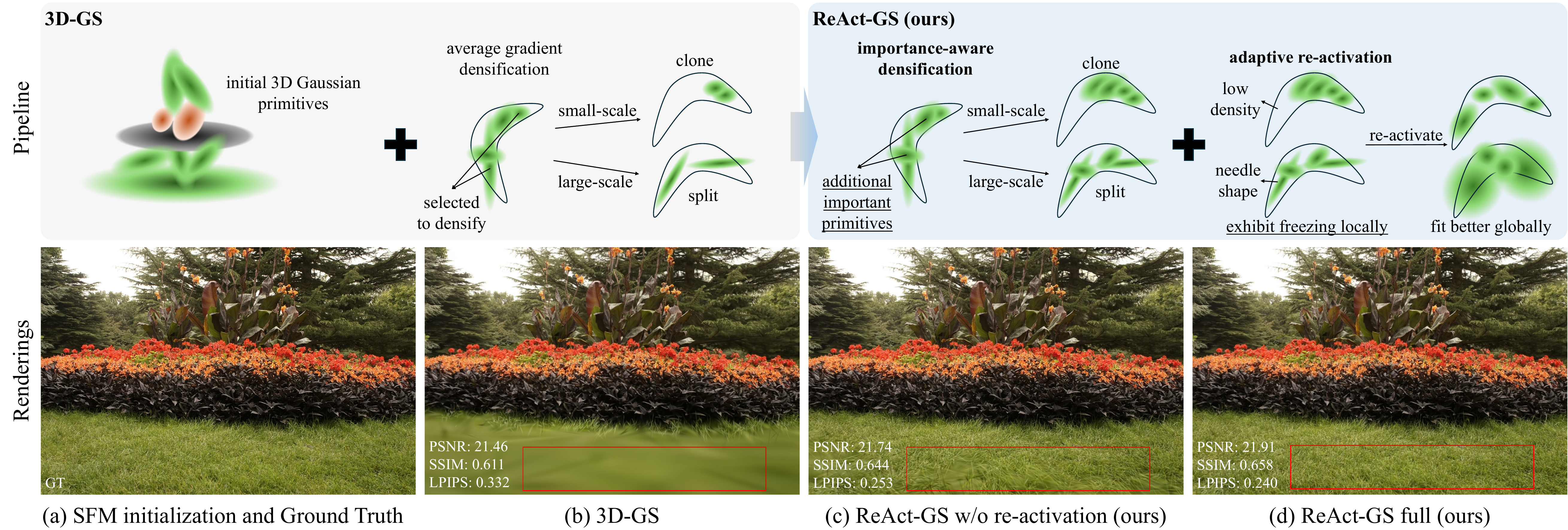}
  \caption{We present ReAct-GS, a method that addresses the over-reconstruction issue in 3D-GS by identifying and resolving two fundamental limitations: gradient magnitude dilution and primitive frozen phenomenon. We introduce a novel importance-aware densification criterion and an adaptive re-activation mechanism that effectively eliminate over-reconstruction artifacts in complex scenes, achieving improved rendering quality while accurately preserving fine details in high-frequency regions. The code will be publicly available at \href{https://react-gs.github.io/}{\textcolor{c1} {https://react-gs.github.io}}.}
  \label{fig:titlefigure}
\end{teaserfigure}

\maketitle

\section{Introduction}
\label{sec:intro}
The advancements in 3D Gaussian Splatting (3D-GS)~\cite{kerbl20233d} have established it as a promising approach for explicit point-based neural representation, demonstrating significant potential across applications from immersive VR/AR to robotics~\cite{zhai2025splatloc, fr_3dgs_vr, lu2024manigaussian}. This representation paradigm enables exceptionally rapid training and real-time rendering while maintaining photorealistic quality. These capabilities stem from 3D-GS's distinctive ability to adaptively refine Gaussian primitives through densification and optimization, offering an optimal balance between computational efficiency and visual fidelity. However, when applied to complex scenarios (e.g., unbounded outdoor or high-frequency texture), 3D-GS exhibits persistent over-reconstruction artifacts. These artifacts manifest as local blurring and loss of fine details, primarily attributed to regions dominated by suboptimally scaled Gaussians (either oversized or needle-shape). Addressing these quality degradation issues has become a critical challenge in advancing 3D-GS technology.

\begin{figure}
    \centering
    \includegraphics[width=\linewidth]{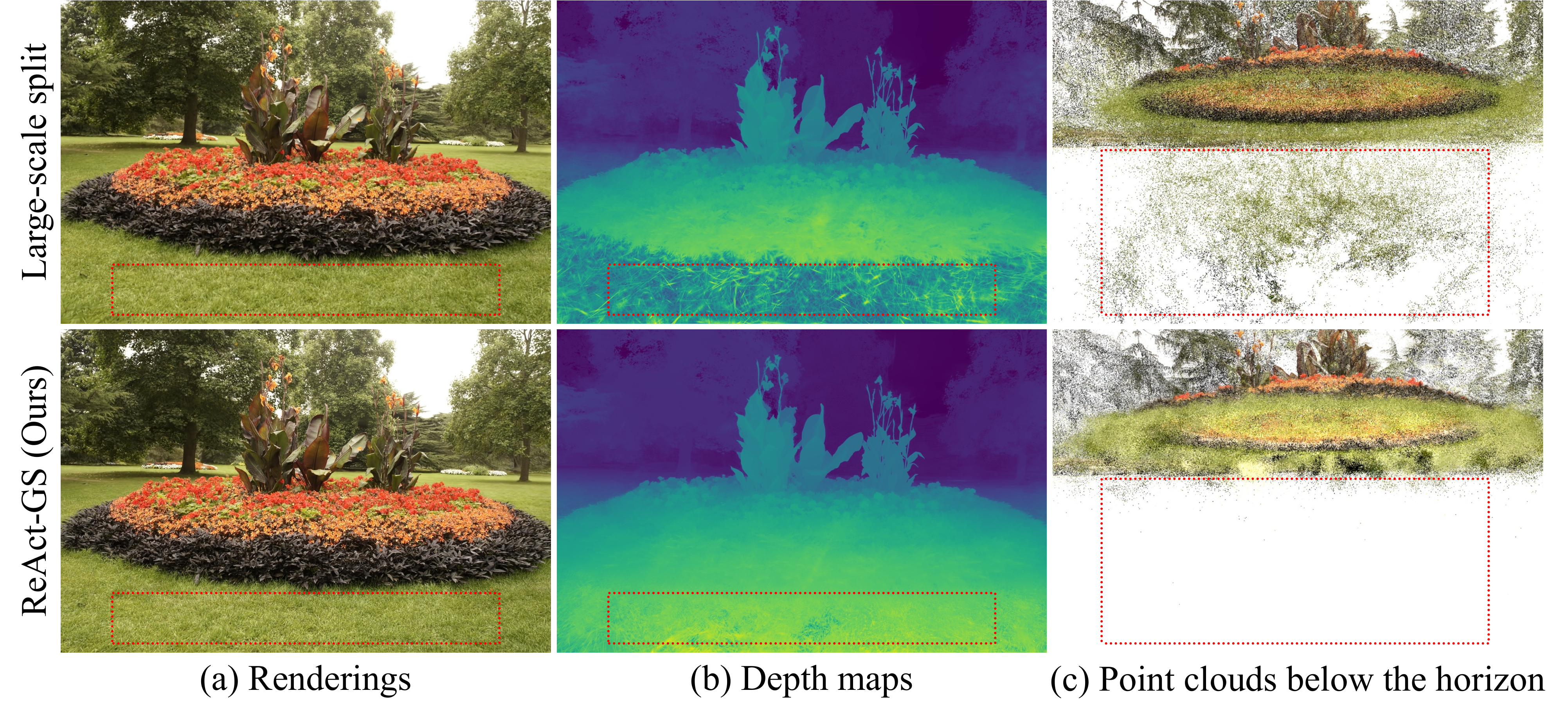}
    \caption{Visualization of renderings, depth maps and point cloud distributions in an unbounded 3D scene. The method Abs-GS~\cite{ye2024absgs}, relying on large-scale splitting strategy (upper), leads to ill-position Gaussian growth in complex regions.}
    \label{fig:largescalesplit}
\end{figure}

Recent research has primarily focused on aggressive splitting strategies to mitigate over-reconstruction in 3D-GS. Notable approaches such as Abs-GS~\cite{ye2024absgs} and Mini-Splatting~\cite{fang2024mini} propose large-scale Gaussian splitting to reduce rendering artifacts caused by oversized primitives. While these methods achieve improved visual quality, they suffer from a critical limitation: excessive splitting leads to premature fragmentation of background primitives and potentially biases the growth of 3D Gaussians toward physically implausible locations as an unintended consequence (see~\Cref{fig:largescalesplit}). 

In this study, we identify two fundamental limitations in the current 3D-GS pipeline that collectively cause over-reconstruction. Through systematic analysis, we first demonstrate that the original average gradient criterion suffers from gradient magnitude dilution. We elucidate that gradient magnitude strongly correlates with a 3D Gaussian's contribution to $\alpha$-blending from corresponding viewpoints. This correlation causes critical growth signals from dominant viewpoints to be suppressed when averaged with weaker gradients from less influential perspectives, thereby inhibiting densification in regions requiring fine detail representation. More critically, we discover the previously overlooked primitive frozen phenomenon, where rapidly converged small-scale and needle-shape 3D Gaussians frequently become trapped in optimization stagnation. Their contracted receptive fields and attenuated gradients prevent effective parameter updates, creating persistent artifacts that cannot be resolved merely through aggressive densification. This frozen phenomenon explains why existing methods focusing solely on densification criterion modifications still fail to fully address over-reconstruction, despite employing more primitives~\cite{zhang2024pixel}.

Built upon these insights, we propose ReAct-GS, a method rooted in the principle of re-activation to address over-reconstruction. To address gradient magnitude dilution, we develop an importance-aware densification criterion that re-activates stalled growth in complex regions by incorporating $\alpha$-blending weights from multiple viewpoints into gradient aggregation during densification. To eliminate the primitive frozen phenomenon, we introduce a novel re-activation mechanism that revitalizes trapped primitives through adaptive parameter perturbations, thereby expanding their perceptual range. While such perturbation might intuitively compromise training stability, our experiments demonstrate that it successfully re-activates frozen primitives for continued optimization, achieving both efficient primitive utilization and effective artifact removal in challenging scenarios.

We evaluate ReAct-GS across multiple public real-world datasets, demonstrating significant improvements in both quantitative metrics and qualitative results. Our method achieves superior performance on standard novel view synthesis metrics while retaining comparable memory consumption to state-of-the-art approaches. As shown in~\Cref{fig:titlefigure} and ~\Cref{fig:largescalesplit}, our approach effectively resolves the over-reconstruction problem while preserving geometrically accurate structures, as evidenced by both rendered images and depth maps. In summary, our contributions are as follows:
\begin{itemize}
    \item We identify and analyze two fundamental limitations in current 3D-GS pipelines: gradient magnitude dilution and primitive frozen phenomenon, which collectively cause over-reconstruction artifacts.
    \item We propose the ReAct-GS with a novel re-activation principle, implemented by importance-aware densification and adaptive re-activation mechanism, to address the limitations.
    \item Extensive experiments demonstrate that our method successfully eliminates over-reconstruction artifacts while achieving superior novel view synthesis quality and geometrically precise reconstruction.
\end{itemize}

\section{Related Works}
\label{sec:relat}
\noindent\textit{\textbf{Gaussian Splatting.}}
3D Gaussian Splatting (3D-GS)~\cite{kerbl20233d}, a representative point-based 3D rendering method, has garnered increasing attention in the novel view synthesis domain. Unlike previous structured implicit radiance fields, e.g., NeRF~\cite{mildenhall2021nerf}, 3D-GS pioneers the reconstruction of 3D scenes using 3D elliptical Gaussian primitives with learnable parameters. Through elaborate optimization strategies coupled with parallelizable splat-style rasterization techniques~\cite{zwicker2001ewa}, 3D-GS has quickly emerged as a protagonist among mainstream 3D neural representations due to its fast training and computational efficiency and real-time high-fidelity rendering capabilities. Based on 3D-GS's advanced characteristics, numerous follow-up studies have explored enhancements and applications, including alias-free rendering~\cite{yu2024mip}, efficient training~\cite{mallick2024taming, hanson2025pup, navaneet2024compgs}, editing and generation~\cite{chen2024gaussianeditor, yi2024gaussiandreamer, cheng2025perspective}, etc.~\cite{rai2025uvgs, zhu2025_loopsplat, lu2024scaffold, fang2024mini}. While extended topics built upon 3D-GS are under active exploration~\cite{wang2024view, zhou2024drivinggaussian, cai2024radiative}, its critical over-reconstruction deficiency~\cite{kerbl20233d} remains unsolved and impedes further deployment in complex real-world scenarios~\cite{rota2024revising}.

\noindent\textit{\textbf{Over-reconstruction and densification.}} 3D-GS typically shows textural degradation in representing high-frequency regions or fine-grained repetitive textures, which is defined as over-reconstruction. Essentially, this issue manifests when covering complex scenes with only a small number of 3D Gaussian primitives that are difficult to optimize~\cite{kerbl20233d}, attributed to sparse point cloud initialization~\cite{schoenberger2016sfm, zhang2024pixel} and the inherent absence of periodicity features in Gaussian function~\cite{zhang2024fregs}. The vanilla Adaptive Density Control (ADC) strategy~\cite{kerbl20233d} addresses this by splitting primitives based on average gradient magnitude under Normalized Device Coordinates (NDC). However, this approach struggles with challenging conditions, particularly unbounded outdoor scenes and tiny details. Recent studies propose various solutions to this limitation. Pixel-GS~\cite{zhang2024pixel} introduces dynamic pixel-aware gradients by incorporating projected pixel numbers as weights to facilitate large-scale primitive splitting, while Abs-GS~\cite{ye2024absgs} addresses gradient collision flaw through absolute operator to encourage similar splitting behavior. Mini-Splatting~\cite{fang2024mini} takes a different approach by employing a threshold to identify and actively split large-scale Gaussians that cause blur in rendering, augmenting the original gradient-based criterion. Additional improvements include residual densification~\cite{lyu2024resgs}, smooth densification~\cite{kheradmand20243d}, and visual consistent split~\cite{feng2024new}, among others~\cite{Kim_2024_CVPR, yuan2025ema, wang2025hda, lee2024oda}.

Building upon prior works, we conduct a thorough analysis of current gradient-based densification schemes and demonstrate that the differential treatment of large-scale primitives in densification adversely affects the final 3D scene representation. Consequently, we propose an advanced densification criterion that incorporates importance-awareness. Furthermore, we identify the primitive frozen phenomenon as another critical factor of over-reconstruction artifacts and introduce an adaptive re-activation mechanism to address these challenges.

\section{Method}\label{sec:metho}
In this section, we first review the background of 3D-GS in~\Cref{subsec:preli}; then, we analyze the gradient magnitude dilution issue in current gradient criterion and propose importance-aware densification in~\Cref{subsec:importance_aware_densification}; finally, we introduce the primitive frozen phenomenon and re-activation mechanism to overcome such limitation and effectively address over-reconstruction artifacts in ~\Cref{subsec:reactivation_mechanism}.

\subsection{Preliminaries}\label{subsec:preli}
\textit{\textbf{3D Gaussian representation.}} 3D-GS~\cite{kerbl20233d} fits a scene by optimizing a set of learnable 3D Gaussian primitives $\{G_i \ | \ i=1, \cdots, N\}$. A 3D Gaussian $G_i$ can be represented in 3D space as:
\begin{equation}
G_i(\bx) = e^{-\frac{1}{2} (\bx-\bmu_i^{3D})^T (\varSigma_i^{3D})^{-1}(\bx-\bmu_i^{3D})},
\label{eq:3DGaussians}
\end{equation}
where $\bmu_i^{3D} \in \RRR^{3 \times 1}$ and $\varSigma_i^{3D} \in \RRR^{3\times3}$ represent the center position and covariance matrix, respectively. Additionally, each Gaussian primitive is also characterized with two extra attributes: opacity $o_i\in\left[0,1\right]$ and color $\bc_i$ in spherical harmonic (SH) coefficients to encode view-dependent color for rendering. During splatting, the 3D Gaussian primitive $G_i$ is projected to 2D screen space by 6-DoF transformation matrix and affine approximation, obtaining $G_i^{2D}$ with $\bmu_i^{2D}$ and $\varSigma^{2D}_i$. With extensive number of primitives splatting onto image plane, 3D-GS utilizes differentiable $\alpha$-blending to render the color for each pixel $\bp$ in the depth order as:
\begin{equation}
        \bc(\bp) = \sum^{N}_{i=1} \bc_{i} \omega_i, \omega_i = \alpha_{i}(\bp) \prod_{j=1}^{i-1}(1-\alpha_{j} (\bp)), \alpha_{i} (\bp) = o_i G_i^{2D}(\bp), 
    \label{eq:alpha_blending}
\end{equation}
where $N$ is the number of Gaussians involved in $\alpha$-blending~\cite{zwicker2001ewa}. These 3D Gaussian parameters are optimized under multi-view supervision through a composite photometric loss $L$.

\noindent\textit{\textbf{Adaptive Density Control (ADC).}} Since Gaussians are initialized from sparse point clouds produced from SfM~\cite{schonberger2016structure}, ADC strategy is designed to densify 3D Gaussian primitives to better represent sparse areas. To address the \textit{over-reconstruction} and \textit{under-reconstruction}, ADC conducts \textit{split} or \textit{clone} operation for large or small Gaussian primitives, respectively. For a 3D Gaussian $G_i$ and its 2D image plane projection center $\bmu_i^{k}=(\bmu_{i,x}^{k}, \bmu_{i,y}^{k})$ under the $k$ viewpoint, the densification will be performed if its average gradient $\nabla_{\bmu_i}L$ on screen space over $M$ viewpoints satisfies
\begin{equation}
    \nabla_{\bmu_i}L = 
    \frac{1}{M}{\sum_{k=1}^{M}\left\|\nabla_{\bmu_i}L^k\right\|} =
    \frac{1}{M}{\sum_{k=1}^{M}\sqrt{\left(\frac{\partial L^k}{\partial\bmu_{i,x}^k}\right)^2 + \left(\frac{\partial L^k}{\partial\bmu_{i,y}^k}\right)^2}},
    \label{eq:vanilla_grdient}
\end{equation}
where $\tau_{pos}$ is a predefined threshold for recognizing large gradient elements, and another hyper-parameter $\tau_{size}$ is set to discriminate large or small primitives by their largest principal scale:
\begin{equation}
            \nabla_{\bmu_i}L > \tau_{pos} \ \ \text{and} \ \left\{  
             \begin{array}{lr}
             \max(s_x, s_y, s_z) > \tau_{size}, \ \ \ \ \text{split}&\\  
             \max(s_x, s_y, s_z) < \tau_{size}, \ \ \ \ \text{clone}&\\     
             \end{array}  
    \right.
\label{eq:vanilla_densification_criteria}
\end{equation}

\subsection{Importance-Aware Densification}\label{subsec:importance_aware_densification}
Despite advancements in addressing over-reconstruction, the unintended artifact in~\Cref{fig:largescalesplit}, resulting from the large-scale 3D Gaussian splitting technique~\cite{ye2024absgs, fang2024mini}, indicates this problem remains unsolved. Through an in-depth analysis of the average gradient densification criterion, we identify a key factor--gradient magnitude dilution--which causes densification to stall when fitting complex regions by overlooking the primitive's importance in $\alpha$-blending rendering. In response to this issue, we propose an importance-aware densification criterion to re-activate the densification process.

\begin{figure}[t]
    \centering
    \includegraphics[width=\linewidth]{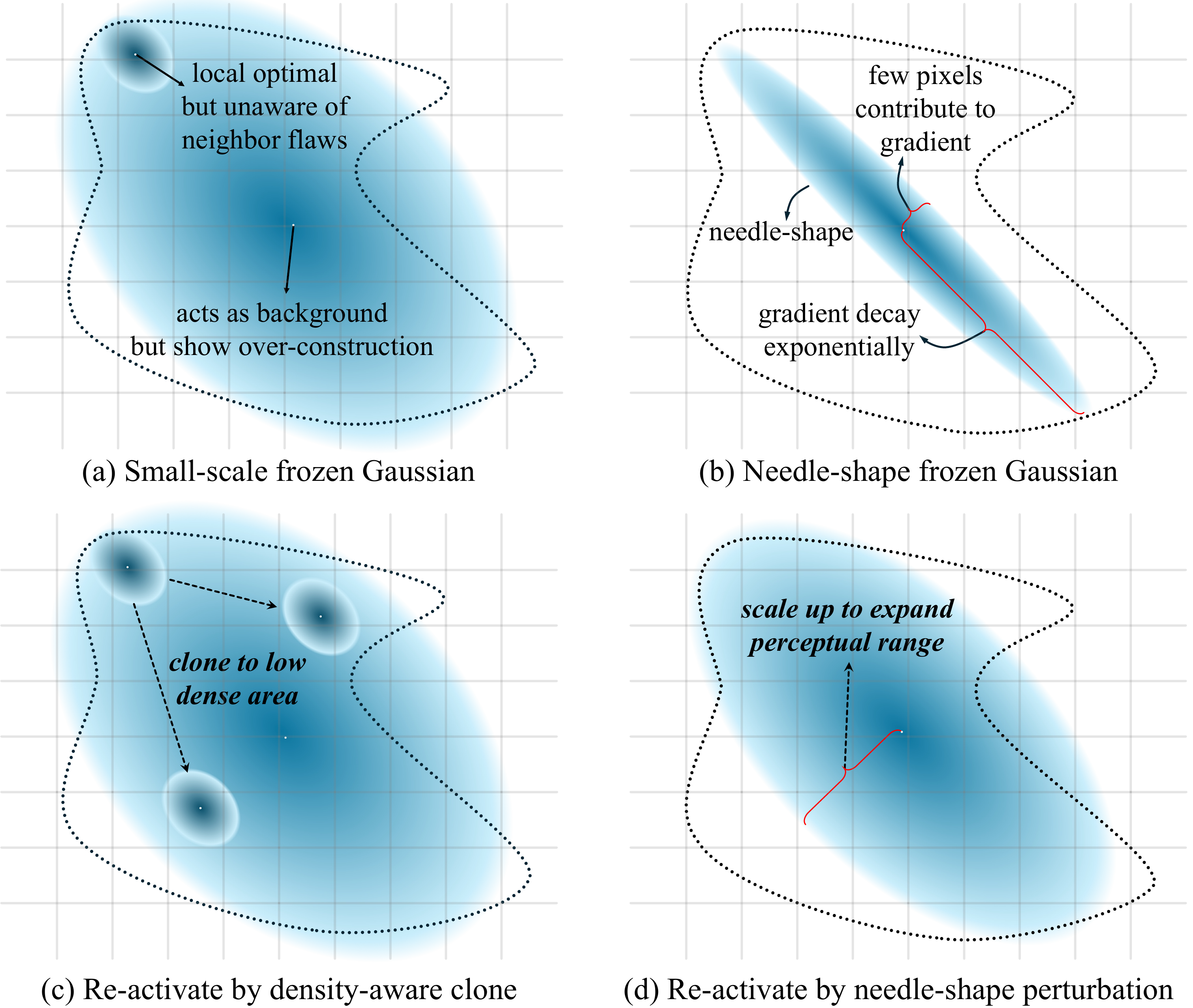}
    \caption{Illustration of primitive frozen phenomenon and our adaptive re-activation mechanism. (a) and (b) depict the underlying factors causing small-scale and needle-shape 3D Gaussians to freeze; (c) and (d) showcase our proposed re-activation strategies for these respective frozen primitives.}
    \label{fig:reactivation}
\end{figure}

For Gaussian $G_i$ with projection center $\bmu_i^k$ on $k$-th viewpoint, we assume that it is splatted onto $m_i^k$ pixels $\{\bp_1, \bp_2, \dots, \bp_{m_i^k}\}$, obtaining the gradient $\nabla_{\bmu_i}L^k$ of \Cref{eq:vanilla_grdient} in NDC space as follows:
\begin{equation}
    \nabla_{\bmu_i}L^k = 
    \sum_{j=0}^{m_i^k}\nabla_{\bmu_i}L_j^k = 
    {\left(
        \sum^{m_i^k}_{j=0}\frac{\partial L_j^k}{\partial \bmu_{i,x}^k},  \ \
        \sum^{m_i^k}_{j=0}\frac{\partial L_j^k}{\partial \bmu_{i,y}^k}
    \right)}.
    \label{eq:none}
\end{equation}
Focusing on the $x$-axis component $\nabla_{\bmu_i}L_j^k$ of the gradient with respect to the $j$-th pixel, we derive the following analytical formulation:
\begin{equation}
    \frac{\partial L_j^k}{\partial \bmu_{i,x}^k} 
    = \frac{\partial L_j^k}{\partial c(\bp_j)}\times \frac{\partial c(\bp_j)}{\partial \alpha_i(\bp_j)}\times \frac{\partial \alpha_i(\bp_j)}{\partial \bmu_{i,x}^k} \\
    \label{eq:ndc_gradient_x_axis}
\end{equation}
The above discussion is simplified to consider only a single channel of the RGB color. The term $\partial L_j^k/\partial c(\bp_j)$ corresponds to the gradient of the loss function employed in the optimization process. For the remaining terms, we derive their detailed expressions according to the $\alpha$-blending formulation presented in~\Cref{eq:alpha_blending}:
\begin{equation}
    \begin{aligned}
    & \frac{\partial c}{\partial \alpha _i} = c_i\prod_{j=1}^{i-1}(1-\alpha_j) + \sum_{l=i+1}^{N} (-c_l\alpha_l\prod_{j=1, j\neq i}^{l-1}(1-\alpha_j)) \\
    & \frac{\partial \alpha _i}{\partial \bmu_{i,x}^k} = o_i\times G_i^{2D}\times g'_{i,x},  \ \ \frac{\partial G_i^{2D}}{\partial \bmu_{i,x}^k} = G_i^{2D}\times g'_{i,x}
    \end{aligned}
    \label{eq:alpha_blending_gradient}
\end{equation}
where $\bp_j$ is omitted for clarity, and $c$ and $\alpha$ represent the corresponding color and alpha values for each Gaussian in depth order. The term $\partial G_i^{2D}/\partial \bmu_{i,x}$ is expressed as $G_i^{2D}\times g'_{i,x}$ where $g'_{i,x}$ is a complex item dependent on the projected 2D covariance matrix of $G_i$ and the distance between $\bp_j$ and $\bmu_i^k$. After combining and simplifying, we derive the following equation (detailed in Appendix):
\begin{equation}
     \frac{\partial c}{\partial \alpha _i}\times \frac{\partial \alpha _i}{\partial \bmu_{i,x}^k} = \omega_i^k \cdot \left\{
        c_i - \sum_{l=i+1}^{N} \left[c_l\cdot \alpha_l\prod_{j=i+1}^{l-1}(1-\alpha_j)\right]
        \right\}\cdot g_{i,x}'
    \label{eq:combination_last_two_term_gradient}
\end{equation}
where $\omega_i^k$ is $G_i$'s rendering weight defined in~\Cref{eq:alpha_blending}. \Cref{eq:combination_last_two_term_gradient} reveals that the gradient magnitude of $G_i$ from the $k$-th viewpoint is positively correlated with its importance in rendering. When $G_i$ is located in a complex area from the $k$-th viewpoint with large $\omega_i^k$ (e.g., positioned earlier in depth order) but has low rendering weights from other views (positioned later in depth order), the average gradient magnitude in~\Cref{eq:vanilla_grdient} can fall below $\tau_{pos}$, stagnating its densification and leading to persistent blurring artifacts. Intuitively, gradients of $G_i$ from viewpoints where it plays a significant role (higher $\omega_i$) should have greater influence in determining whether densification is needed for optimal representation. However, this limitation severely hinders the growth of model capacity to reconstruct complex scenes--an effect we term as gradient magnitude dilution and identify as a fundamental cause of the over-reconstruction problem. 

Based on theoretical analysis, we propose a novel and principled approach to re-activate stalled densification, termed importance-aware densification. Considering the densification of $G_i$ after optimization through $M_i$ views with $\{m_i^k | k\in \{1,\dots, M_i\}\}$ pixels splatted for each view, we integrate the rendering importance weights $\{\omega_{i,j}^{k} | j\in\{1,\dots,m_k\}\}$ into the $\nabla_{\bmu_i}L$ calculation as:
\begin{equation}
    \nabla_{\bmu_i}L = \frac{\sum_{k=1}^{M}\omega_i^k\sqrt{\left(\frac{\partial L^k}{\partial\bmu_{i,x}^k}\right)^2 + \left(\frac{\partial L^k}{\partial\bmu_{i,y}^k}\right)^2}}{\sum_{k=1}^M \omega_i^k }, \ \ \omega_i^k = \frac{\sum_{j=1}^{m_i^k}\omega_{i,j}^{k}}{m_i^k}
    \label{eq:importance_aware_grdient}
\end{equation}
where the $\sum_{j=0}^{m_i^k}\omega_{i,j}^{k}$ can be accumulated during the forward pass without introducing additional computational overhead. Crucially, we normalize the accumulated $\omega_{i,j}^{k}$ values by the number of pixels $m_i^k$ that $G_i$ splats onto each viewpoint, which effectively mitigates densification bias towards Gaussians with larger projected areas. This normalization strategy is essential for preventing inappropriate splitting of large-scale Gaussians, as demonstrated in ~\Cref{fig:largescalesplit}.

\sloppy
Our comprehensive experiments demonstrate that the proposed importance-aware densification criterion consistently outperforms existing approaches in high-frequency texture representation by systematically cultivating critical primitives to restore over-reconstruction regions, as thoroughly analyzed in~\Cref{sec:exper}.

\begin{figure}[t]
    \centering
    \includegraphics[width=\linewidth]{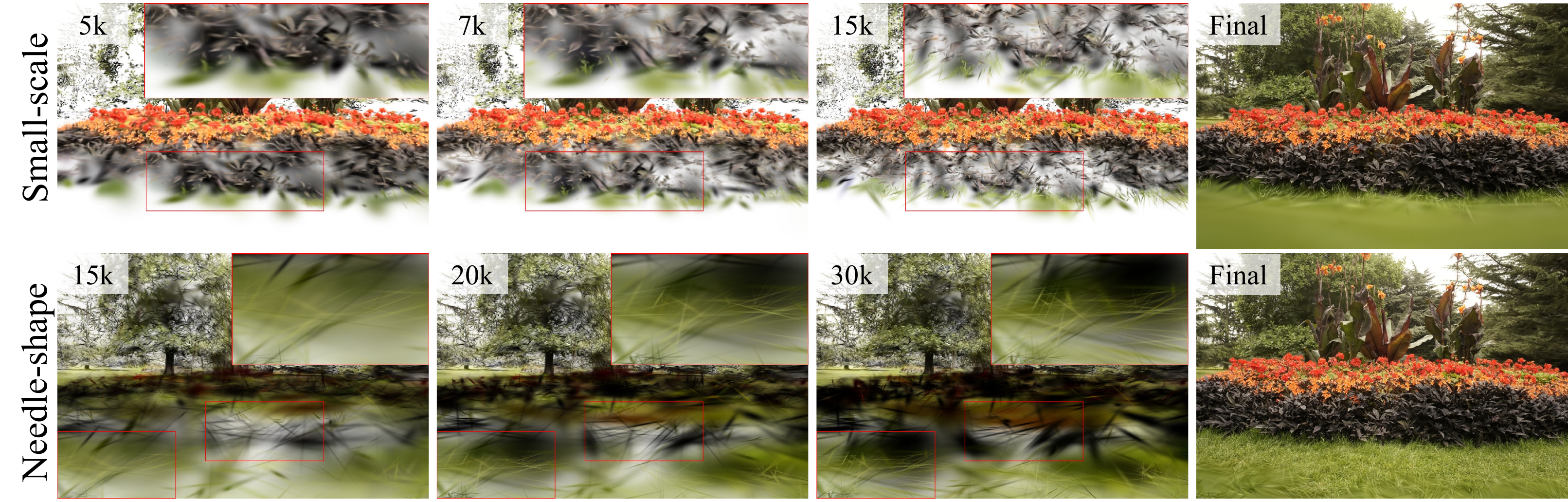}
    \caption{Visualization of primitive frozen phenomenon across optimization stages. \textit{Top}: Small-scale 3D Gaussians freeze locally and barely exhibit growth to relieve blurry during densification; \textit{Bottom}: Needle-shape primitives persist and cannot be mitigated in post-densification optimization.}
    \label{fig:reactivation_exp}
\end{figure}

\subsection{Re-Activation Mechanism}\label{subsec:reactivation_mechanism}
Although enhanced densification strategies demonstrate substantial improvement, they remain inadequate for completely addressing all over-reconstruction artifacts. Through rigorous analysis, we identify another fundamental limitation: the primitive frozen phenomenon, wherein established Gaussians become resistant to subsequent optimization refinement.

\begin{table*}[htbp]
\caption{
Quantitative results on Mip-NeRF 360\cite{barron2021mip}, Tanks \& Temples\cite{knapitsch2017tanks}, and Deep Blending\cite{hedman2018deep}. $^\dagger$ indicates that the metrics are directly sourced from the original 3D-GS report~\cite{kerbl20233d}. The scores are highlighted as: \colorbox{red!40}{1st-}, \colorbox{orange!40}{2nd-}, and \colorbox{yellow!40}{3rd-} best performances, respectively. The MB and GB represent Megabyte and Gigabyte in storage.
}
\label{tab:main_results}
\resizebox{\linewidth}{!}{
    \begin{tabular}{l|cccc|cccc|cccc}
    \toprule[1.4pt]
    Datasets & \multicolumn{4}{c|}{Mip-NeRF 360} & \multicolumn{4}{c|}{Tanks \& Temples} & \multicolumn{4}{c}{Deep Blending} \\ 
    \midrule[0.6pt]
    Methods & PSNR $\uparrow$ & SSIM $\uparrow$ & LPIPS $\downarrow$ & Mem & PSNR $\uparrow$ & SSIM $\uparrow$ & LPIPS $\downarrow$ & Mem & PSNR $\uparrow$ & SSIM $\uparrow$ & LPIPS $\downarrow$ & Mem \\ 
    \midrule[1.0pt]
    Plenoxels $^\dagger$~\cite{fridovich2022plenoxels} & 23.08  & 0.626  &0.463  & 2.1GB & 21.08 & 0.719 & 0.379 & 2.3GB & 23.06 & 0.795 & 0.510 & 2.7GB \\
    INGP-Big $^\dagger$~\cite{muller2022instant} & 25.59 & 0.699 & 0.331  & 48MB  & 21.92 & 0.745 & 0.305 & 48MB  & 24.96 & 0.817 & 0.390  & 48MB \\
    Mip-NeRF360 $^\dagger$~\cite{barron2021mip} & \cellcolor{yellow!40} 27.69 & 0.792 & 0.237 & 8.6MB & 22.22 & 0.759 & 0.257 & 8.6MB & 29.40 & 0.901 & 0.245 & 8.6MB \\
    \midrule[0.6pt]
    3D-GS $^\dagger$~\cite{kerbl20233d} & 27.21 & 0.815 & 0.214 & 734MB & 23.14 & 0.841 & 0.183 & 411MB & 29.41 & 0.903 & 0.243 & 676MB \\
    3D-GS~\cite{kerbl20233d} & 27.29 & 0.808 & 0.225 & 793MB & \cellcolor{yellow!40}23.61 & 0.853 & 0.173 & 347MB & \cellcolor{yellow!40}29.41 & \cellcolor{yellow!40}0.907 & 0.195 & 515MB \\
    Blur-split~\cite{fang2024mini} & 27.11 & 0.810 & 0.208  & 634MB & 22.98 & 0.844 & \cellcolor{yellow!40}0.158 & 638MB & 29.09 & 0.898 & 0.194 & 950MB \\ 
    Abs-GS~\cite{ye2024absgs} & 27.27 & \cellcolor{yellow!40} 0.816 & \cellcolor{yellow!40}0.196 & 678MB & 23.39 & \cellcolor{yellow!40}0.857 & 0.161 & 417MB & 29.36 & 0.907 & \cellcolor{red!40}0.190 & 587MB \\ 
    Pixel-GS~\cite{zhang2024pixel} & \cellcolor{orange!40}27.72 & \cellcolor{orange!40} 0.832 & \cellcolor{orange!40}0.178 & 1145MB & \cellcolor{orange!40}23.75 & \cellcolor{orange!40}0.862 & \cellcolor{orange!40}0.149 & 708MB & \cellcolor{orange!40}29.49 & \cellcolor{orange!40}0.907 & \cellcolor{yellow!40}0.191 & 760MB \\ 
    \midrule[0.6pt]
    ReAct-GS (ours) & \cellcolor{red!40} 27.79 & \cellcolor{red!40} 0.835 & \cellcolor{red!40} 0.176 & 805MB & \cellcolor{red!40} 24.06 & \cellcolor{red!40} 0.865 & \cellcolor{red!40} 0.145 & 787MB & \cellcolor{red!40} 29.66 & \cellcolor{red!40} 0.908 & \cellcolor{orange!40} 0.191 & 585MB \\
     \bottomrule[1.1pt]
    \end{tabular}
}
\end{table*}

\noindent\textit{\textbf{Primitive frozen phenomenon.}}
During 3D-GS training, 3D Gaussian primitives rapidly converge to fit local textures within a few iterations. However, we observe that this premature convergence impedes continuous optimization of specific primitives. Combining our analysis from~\Cref{eq:combination_last_two_term_gradient} with the $\alpha$-blending formulation in~\Cref{eq:alpha_blending}, the gradient magnitude of primitives is related to $\alpha_{i,j}^k$, which can be decomposed as:
\begin{equation}
    \label{alpha_i}
    \alpha_{i,j}^k=o_i\times e^{-\frac{1}{2}(\bp_j - \bmu_i^k)^T\varSigma _{2D}^{i,k}(\bp_j - \bmu_i^k)}
\end{equation}
This formulation reveals that gradient magnitude $\nabla_{\bmu_i}L_j^k$ diminishes exponentially with increasing distance between pixel $\bp_j$ and projection center $\bmu_i^k$.
As noted in Pixel-GS~\cite{zhang2024pixel}, only pixels in close proximity to the projection center $\bmu_i^k$ contribute substantially to the gradient of these Gaussian primitives. We propose that this intrinsic characteristic fundamentally restricts small-scale and needle-shape primitives from optimization beyond their initial convergence state.

In~\Cref{fig:reactivation}(a), a small-scale primitive converges to fit local content with minimal pixel contribution to its gradients. Due to its limited spatial perception, this primitive cannot adequately expand to cover blurry regions, perpetuating over-reconstruction artifacts. Similarly, the needle-shape primitive illustrated in~\Cref{fig:reactivation}(b) exhibits constrained optimization: its short axis is influenced by only a few adjacent pixels while gradient magnitude exponentially decays along the long axis, severely limiting its capacity to deform under standard optimization procedures. 

We empirically visualize the evolution of 3D Gaussians to verify our analysis, as shown in~\Cref{fig:reactivation_exp}. We render small-scale and needle-shape Gaussians in densification and post-densification stages, respectively. Theoretically, small-scale Gaussians should spread to cover blurry regions and represent fine details. However, as evident in the upper section of~\Cref{fig:reactivation_exp}, the distribution of these Gaussians exhibits minimal migration toward over-reconstruction regions between 5k and 15k iterations, demonstrating localized freezing behavior. Similarly, needle-shape Gaussians remain virtually unchanged from 15k iterations onward. The lower section of~\Cref{fig:reactivation_exp} demonstrates that these needle-shape artifacts 
emerge at the post-densification stage's commencement and persist throughout training completion, indicating the inability of existing optimization methods to resolve these elongated structures.

To revitalize these frozen primitives for effective optimization, we propose a re-activation mechanism consisting of Density-Guided Clone and Needle-Shape Perturbation.

\noindent\textit{\textbf{Density-Guided Clone.}}
In contrast to exact primitive replication prevalent in conventional densification, we propose strategically relocating cloned primitives based on local density to populate sparse regions, as illustrated in~\Cref{fig:reactivation}(c). For a 3D Gaussian $G_i$ with $\nabla_{\bmu_i}L$ exceeding the densification threshold, we quantify its local density $d_i$ through the K-nearest-neighbors algorithm:
\begin{equation}
    d_i = \frac{1}{K}\sum_{G_j \in N_K(G_i)}\|\bmu_i - \bmu_j\|
\end{equation}
where $\bmu_j$ represents the center of $G_j$, and $K=3$ balances accuracy and computational efficiency. Utilizing this density metric, we position the cloned primitive $G_i'$ by sampling from the distribution $\mathcal{N}(\bmu_i, d_i\cdot\bI_3)$. This density-guided approach confers sufficient optimization momentum to primitives in sparse regions, enabling them to overcome local optima and respond to neighboring gradients. The mechanism facilitates small Gaussians' migration toward over-reconstruction areas while negligibly perturbing primitives already situated within dense clusters. Our experimental results in~\Cref{fig:ablationflowers} demonstrate the enhanced detail reconstruction achieved through this approach, particularly when collaborating with our proposed importance-aware densification.

\begin{figure*}
    \centering
    \includegraphics[width=\linewidth]{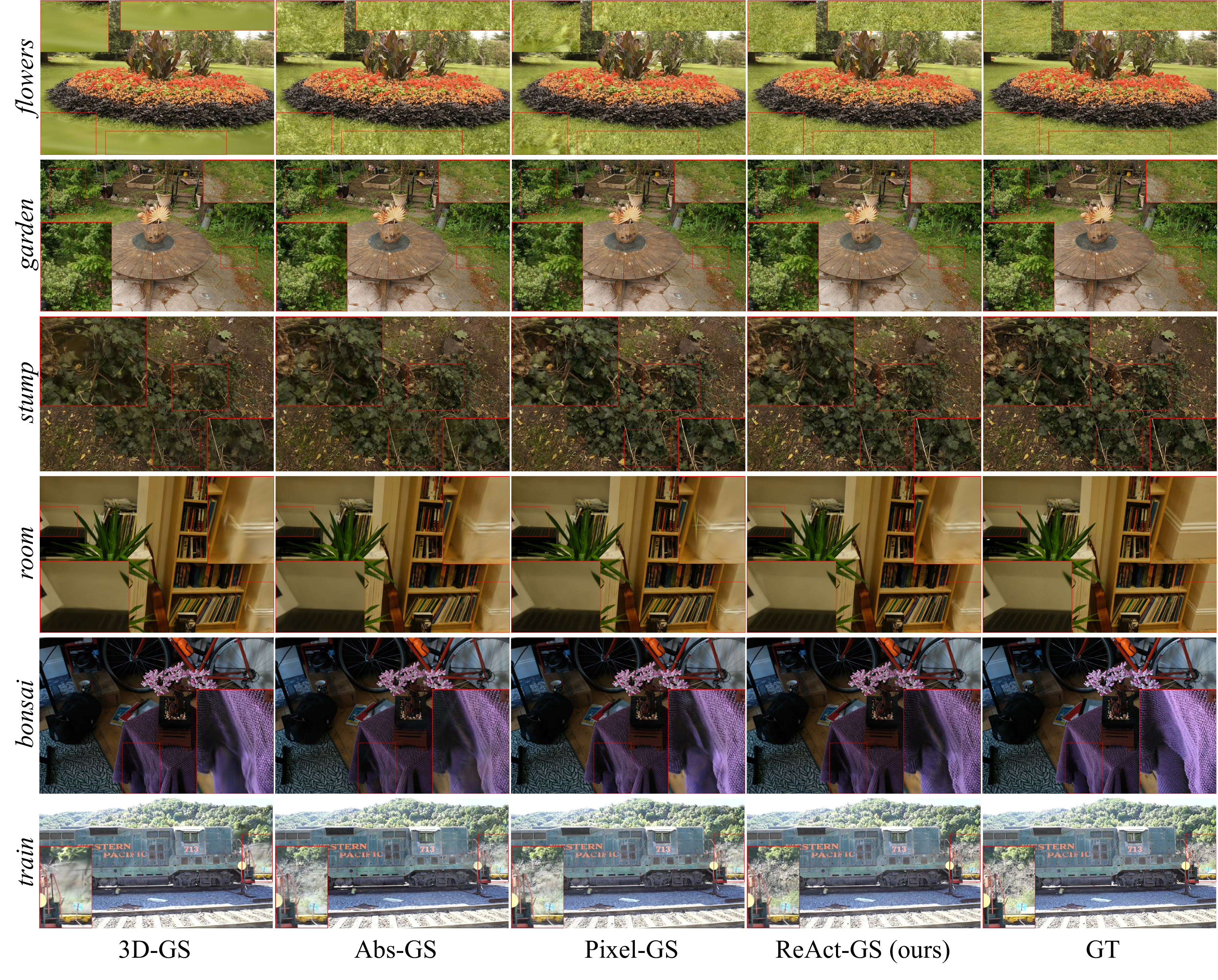}
    \caption{Qualitative comparisons of different methods on scenes from the Mip-NeRF 360 and Tanks \& Temples datasets. Close-up views highlight the challenging areas with high-frequency details, where over-reconstruction is particularly pronounced.}
    \label{fig:main_visualization}
\end{figure*}

\noindent\textit{\textbf{Needle-Shape Perturbation.}}
While the aforementioned techniques deliver satisfactory performance in complex scene reconstruction, needle-shape primitives continue to introduce visual artifacts in challenging corner-case renderings. These elongated Gaussians typically do not satisfy clone criteria due to their dominant principal axis. We therefore propose perturbing their shape to expand their perceptual range, as illustrated in~\Cref{fig:reactivation}(d). Specifically, for 3D Gaussians $G_i$ with scale values $s_i = (s_i^1, s_i^2, s_i^3)$, we identify needle-shape primitives using the criterion:
\begin{equation}
    G_{ns}=\left\{G_i|\frac{\max\{s_i\}}{\sum s_i}>\tau_{ns}, i\in\{1,\dots, N\}\right\}
\end{equation}
where $\tau_{ns}$ is a threshold set to 0.8 in our implementation. For these identified primitives, we perturb their shape by magnifying the shorter principal axes by a factor $\frac{1}{2}s_{deg}$ where $s_{deg}=\max\{s_i\}/\text{mid}\{s_i\}$. This shape perturbation introduces additional neighboring pixels with significant gradient contribution, facilitating needle-shape Gaussian optimization. Analogous to opacity reset in conventional 3D-GS, we apply needle-shape perturbation periodically at intervals of 3k iterations to minimize disruption to the optimization process. Our experiments demonstrate that this approach effectively eliminates most needle-like artifacts, particularly in rendering corner regions, as detailed in~\Cref{subsec:main_results}.

\section{Experiments}
\label{sec:exper}
\begin{figure*}
    \centering
    \includegraphics[width=\linewidth]{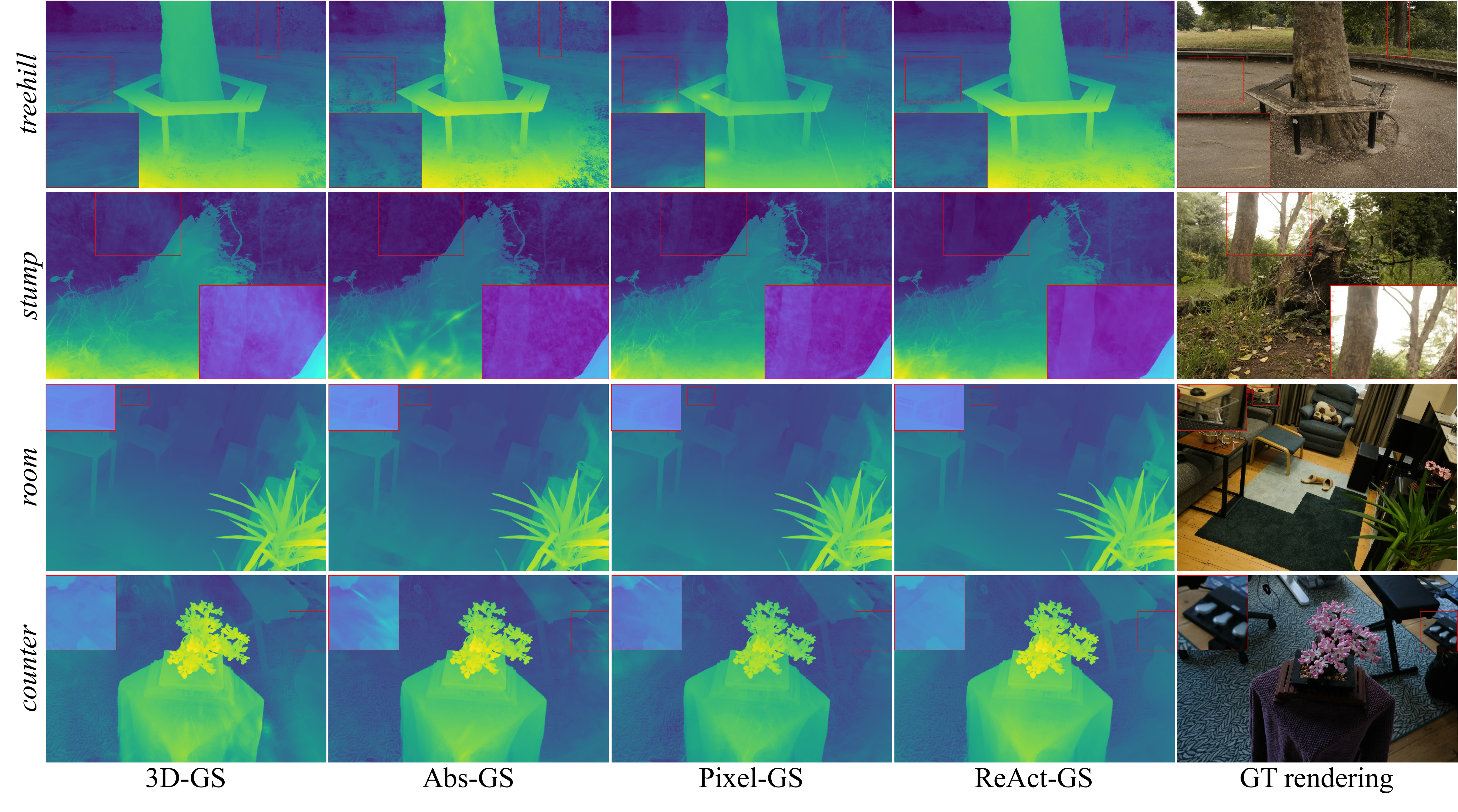}
    \caption{Comparison of depth map renderings across different methods on indoor/outdoor scenes from the Mip-NeRF 360 dataset. Enlarged views highlight the challenging areas where geometric consistency and coherence are crucial.}
    \label{fig:main_depth}
\end{figure*}

\subsection{Setup}
\textit{\textbf{Datasets.}}
Following the 3D-GS~\cite{kerbl20233d} evaluation protocol, we assess ReAct-GS on 13 diverse real-world scenes spanning indoor enclosed spaces and outdoor areas. Specifically, we employ: (1) all 9 indoor/outdoor scenes from Mip-NeRF360~\cite{barron2021mip}, (2) two additional outdoor environments (Train and Truck) from~\cite{knapitsch2017tanks}, and (3) two indoor scenes (drjohnson and playroom) from~\cite{hedman2018deep}. This selection ensures a thorough evaluation across varied settings while maintaining consistency with prior works.

\noindent\textit{\textbf{Baselines.}}
We perform comprehensive comparisons against advanced NeRF-based methods~\cite{fridovich2022plenoxels, muller2022instant, barron2021mip} and recent Gaussian Splatting methods (3D-GS~\cite{kerbl20233d} and its variants designed to address over-reconstruction issues~\cite{fang2024mini, ye2024absgs, zhang2024pixel}). For NeRF-based approaches, we utilize the quantitative metrics from the original publications~\cite{kerbl20233d}. For GS-based approaches, we use official implementations with consistent configurations, disabling auxiliary strategies that might affect fair comparison. Regarding Mini-Splatting~\cite{fang2024mini}, we focus on its core Blur-split module (which targets over-reconstruction mitigation) while omitting other components (depth-reinitialize and pruning), labeled as "Blur-split" in our results for clarity.

\noindent\textit{\textbf{Implementation details.}}
All experiments were conducted on an NVIDIA RTX-3090 GPU with 24 GB of memory. Following standard Gaussian splatting protocols, we adopt the progressive densification schedule, densifying the Gaussian primitives from 500 to 15k at 100-iteration intervals, with training termination at 30k iterations. We maintain the original 3D-GS evaluation with consecutive 8-frame training segments and fixed test frames. Our evaluation reports standard novel view synthesis metrics (PSNR, SSIM, LPIPS) along with memory consumption for well-optimized Gaussian parameter storage. To address over-reconstruction, ReAct-GS employs importance-aware densification during both clone and split operations. This approach inherently amplifies gradient magnitudes during optimization, necessitating an adjusted gradient threshold ($\tau_{pos}=3e^{-4}$) to ensure memory constraints and stable training. For baseline comparisons, we faithfully reproduce their original configurations: $\tau_{pos}=4e^{-4}$ for Abs-GS split operation and $\tau_{pos}=2e^{-4}$ for other methods, as specified in their respective publications.

\subsection{Main Results}\label{subsec:main_results}
\textit{\textbf{Quantitative results.}}
The quantitative comparison across various methods is presented in~\Cref{tab:main_results}. It is noteworthy that ReAct-GS demonstrates consistent superiority across all evaluation metrics, validating the effectiveness of our proposed importance-aware densification and novel re-activation mechanism. While ReAct-GS maintains slightly higher parameter counts (increased memory for storage) than methods focusing exclusively on splitting large-scale primitives (Blur-split and Abs-GS), our method eliminates the depth inconsistency artifacts (shown in~\Cref{fig:largescalesplit}) that plague split-intensive approaches, which is a crucial factor for 3D reconstruction. Notably, ReAct-GS achieves these results with fewer primitives than Pixel-GS, while simultaneously delivering superior performance in both over-reconstruction mitigation and fine detail preservation, demonstrating ReAct-GS's exceptional efficiency in primitive utilization. 

\noindent\textit{\textbf{Qualitative results.}}
In~\Cref{fig:main_visualization}, we compare novel view renderings with state-of-the-art approaches. In addition to significantly reducing the over-reconstruction blur and improving overall visual quality across all scenes, our method also excels in reconstructing fine-grained textures in challenging regions, such as grassy areas at the frame's corners (\textit{flowers}), dense foliage inside the tree stump (\textit{stump}), and object edges (\textit{room}). To further assess geometric reconstruction capability,~\Cref{fig:main_depth} presents depth maps rendered with the official $\alpha$-blending variant~\cite{kerbl20233d}. ReAct-GS produces depth-consistent reconstructions that comply with real-world geometry without spurious floaters, even in complex scenarios like uneven gravel ground (\textit{treehill}). Moreover, ReAct-GS outperforms all competitors in distant object reconstruction, as evidenced by precise depth outlines (e.g., geo-grid in \textit{room}). These improvements stem from our two key designs: (1) The \textit{importance-aware densification} strategically grows primitives close to object surfaces where needed, enhancing detail without overfitting or disrupting Gaussian distributions; (2) our \textit{re-activation mechanism} mitigates over-reconstruction and needle artifacts by recycling frozen Gaussians for global optimization, avoiding parameter inflation while encouraging uniform primitive distribution, thus improving efficiency.

\subsection{Ablation Study}\label{subsec:ablation}
\begin{figure}
    \centering
    \includegraphics[width=\linewidth, height=\linewidth]{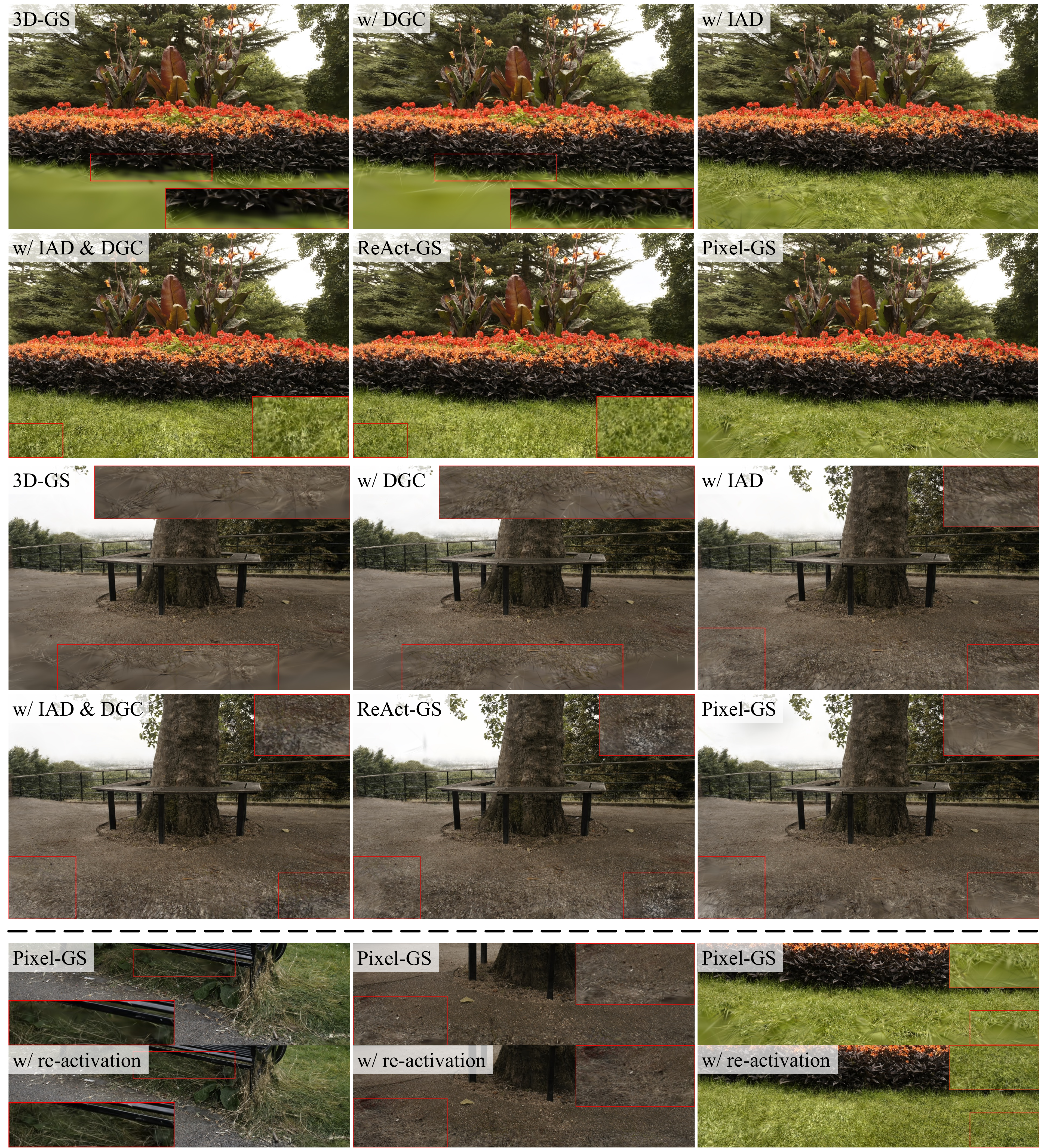}
    \caption{Visualization of ablation experiments. \textit{Top}: Visual improvements from each proposed module on challenging outdoor scenes. \textit{Bottom}: Performance enhancement of Pixel-GS when integrated with our re-activation mechanism.}
    \label{fig:ablationflowers}
\end{figure}

In this section, we rigorously evaluate the efficacy of our proposed modules through ablation experiments with quantitative results presented in~\Cref{tab:ablation}. Built upon the vanilla 3D-GS baseline, we explore four key component combinations. The results demonstrate that our importance-aware densification drives substantial rendering quality improvements (+0.36 dB PSNR avg.) by reviving stalled 3D Gaussian growth to mitigate the persistent over-reconstruction issue. Additionally, the re-activation mechanism effectively resolves the remaining artifacts in challenging regions that existing methods fail to address. Meanwhile, the density-guided clone also operates as a standalone enhancement, improving detail representation even on baseline 3D-GS. Complementing these metrics,~\Cref{fig:ablationflowers} provides visual evidence of each module's necessity, particularly in texture-rich areas where other approaches struggle. The progressive improvements validate our hierarchical design: while importance-aware densification establishes the foundation, re-activation delivers critical refinements for comprehensive scene representation.

To further validate the generalizability of our approach, we investigate the integration potential of our re-activation mechanism with existing methods. Specifically, we select Pixel-GS, which represents the state-of-the-art baseline with its own densification criterion yet still exhibiting over-reconstruction in challenging regions (as shown in the lower part of~\Cref{fig:ablationflowers}). Remarkably, when augmented with our re-activation module, Pixel-GS demonstrates complete elimination of its persistent artifacts, as visually confirmed in~\Cref{fig:ablationflowers}. These improvements are further quantified in the last row of~\Cref{tab:ablation}, where the hybrid approach achieves superior metrics compared to standalone Pixel-GS. These results underscore that our re-activation mechanism maintains effectiveness even when transferred to alternative architectures, and that frozen primitive remains a key factor in leading to over-reconstruction even when advanced densification is employed.

\begin{table}[t]
\caption{Ablation study on the Mip-NeRF 360 dataset. Results show average evaluation metrics, where IAD, DGC, and NSP refer to importance-aware densification, density-guided clone, and needle-shape perturbation, respectively.}
\resizebox{1.0\linewidth}{!}{
 \centering
 \begin{tabular}{l|c c c c}
 \toprule[1.4pt]
      & PSNR $\uparrow$ & SSIM $\uparrow$ & LPIPS $\downarrow$ & Mem \\ 
 \midrule[0.6pt]
    Baseline (3D-GS) & 27.29 & 0.815 & 0.225 & 793MB \\
    \ \ \ + IAD & 27.65 & 0.827 & 0.181 & 739MB \\
    \ \ \ + DGC & 27.33 & 0.815 & 0.221 & 832MB\\
    \ \ \ + IAD + NSP & 27.67 & 0.828 & 0.181 & 766MB\\
    \ \ \ + IAD + DGC & \cellcolor{orange!40} 27.76 & \cellcolor{yellow!40} 0.832 &  \cellcolor{orange!40}0.179 & 810MB \\
    Full-equipped (ReAct-GS) & \cellcolor{red!40} 27.79 & \cellcolor{red!40} 0.835 & \cellcolor{red!40} 0.176 & 805MB \\
\midrule[0.6pt]
    Pixel-GS~\cite{zhang2024pixel} + re-activation & \cellcolor{yellow!40}27.74 & \cellcolor{orange!40}0.833 & \cellcolor{yellow!40} 0.180 & 1191MB \\
 \bottomrule[1.1pt]
 \end{tabular}
}
\label{tab:ablation}
\end{table}

\subsection{Efficiency Analysis}\label{subsec:analysis}
Remarkably, ReAct-GS maintains excellent parameter efficiency while enhancing performance. As demonstrated in~\Cref{tab:main_results}, our approach yields significant quality enhancements without significant parameter expansion. On the Mip-NeRF 360 dataset, our method reduces parameters by 29.7\% (avg. training time of 33 mins on vanilla 3D-GS engine) while outperforming Pixel-GS (avg. training time of 41 mins) both quantitatively and qualitatively. Notably, Pixel-GS represents the state-of-the-art method with minimal geometric inconsistency artifacts. Although our approach requires marginally increased computational resources compared to 3D-GS (avg. 26 mins for training) and Abs-GS (avg. 27 mins for training), we effectively resolve the over-reconstruction phenomenon inherent in 3D-GS and deliver more precise geometric reconstruction than Abs-GS, which are critical prerequisites for practical 3D reconstruction applications. Furthermore, the computational costs could be reduced through prospective pruning and acceleration techniques once reconstruction quality ceases to be a limiting factor. We attribute our efficiency advancement to: (1) targeted densification, wherein our importance-aware criterion strategically densifies 3D Gaussians in critical over-reconstructed regions for visual quality (corroborated by~\Cref{tab:ablation}), thus avoiding superfluous additions of minimally contributing primitives; (2) primitive recycling, where our re-activation mechanism repurposes frozen Gaussians for beneficial optimization rather than indefinitely introducing new elements.

\section{Conclusion}\label{sec:concl}
\sloppy
In this work, we identify the critical underlying factors of over-reconstruction in 3D-GS, particularly the deficiencies of gradient magnitude dilution in current densification strategies and the previously overlooked primitive frozen phenomenon. These limitations lead to persistent artifacts such as blurry textures and needle-shape distortions, even when improved densification criteria are applied. To enhance fine-detail reconstruction in complex scenes, we introduce ReAct-GS equipped with two key modules: importance-aware densification and an adaptive re-activation mechanism (comprising density-guided clone and needle-shape perturbation) to re-activate stalled primitive growth and frozen primitives for continual global optimization. Benefiting from theoretical analysis, ReAct-GS effectively combats over-reconstruction while inherently preventing geometrical inconsistencies. Comprehensive experiments, including both quantitative and qualitative evaluations alongside thorough ablation studies, demonstrate our approach's significant improvements in addressing over-reconstruction. Furthermore, ReAct-GS maintains high parameter efficiency and training speed while exhibiting strong transferability, consistently boosting performance when integrated with other 3D-GS variants.

\begin{acks}
This research was supported by the Theme-based Research Scheme (TRS) project T45-701/22-R of the Research Grants Council (RGC), Hong Kong SAR. We thank all anonymous reviewers for their constructive feedback to improve our paper.
\end{acks}

\bibliographystyle{ACM-Reference-Format}
\balance
\bibliography{sample-base}

\newpage
\appendix

\section{PRILIMINARIES--ADC VARIANTS.}
The vanilla ADC is argued to be insufficient to break through over-reconstruction dilemma and corresponding variants are promoted. 

First, the Mini-Splatting introduces \textit{blur-split} approach to direct split 3D Gaussians with large influential projected area, i.e., $G_{\text{blur}}$, apart from the original gradient-based densification, shown as follows:
\begin{equation}
    G_{\text{blur}}=\{G_i | S_i > \tau_{blur}, i \in (1,\dots, N)\}
    \label{eq:blur_split}
\end{equation}
where $\tau_{blur}$ is a hyper-parameters and $S_i$ is the number of pixels where $G_i$ is of largest weight in alpha blending. 

From another aspect, the pixel-aware weighted average gradients from different views is published in \textit{Pixel-GS} as follows:
\begin{equation}
    \nabla{\bmu_i}L_{\text{Pixel-GS}} = \frac{\sum_{k=1}^{M} m_i^k \cdot f(i,k) \cdot \sqrt{\left(\frac{\partial L^k}{\partial\bmu_{i,x}^k}\right)^2 + \left(\frac{\partial L^k}{\partial\bmu_{i,y}^k}\right)^2}}{\sum_{k=1}^{M}m_i^k}
    \label{eq:pixel_gradient}
\end{equation}
where $m_i^k$ is the number of projected pixels of $G_i$ under $k$-th viewpoint and $f(i,k)$ is a factor to suppress floaters close to camera. The $\nabla{\bmu_i}L_{\text{pixel}}$ magnifies the split possibility of large scale Gaussians initialized within sparse regions. 

Third, the gradient collision phenomenon is discovered by \textit{Abs-GS}. Focused on $x$-axis component of $\nabla{\bmu_i}L$, the $\partial L^k / \partial\bmu_{i,x}^k$ is sum of each independent gradient with respect to loss on pixel $j$ affected by $G_i$. As shown at left part of~\Cref{eq:absgs_gradients}, the hetero-directional gradients cancel each other and reduce the final magnitude when $m_i^k$ goes larger, which hinders the large scale Gaussians densification. Subsequently, the homo-directional gradient is proposed as shown at right part as follows:
\begin{equation}
    \frac{\partial L^k}{\partial\bmu_{i,x}^k} = \sum^{m_i^k}_{j=1}\frac{\partial L_j^k}{\partial \bmu_{i,x}^k} \ \ \rightarrow \ \ \left(\frac{\partial L^k}{\partial\bmu_{i,x}^k}\right)_{\text{Abs-GS}} = \sum^{m_i^k}_{j=1}\left|\frac{\partial L_j^k}{\partial \bmu_{i,x}^k}\right|
    \label{eq:absgs_gradients}
\end{equation}
The same operation is performed for $y$-axis as well. With the absolute operation on sub-gradients of each pixel $j$, the collision phenomenon is eliminated to restore split on large primitives.

\section{GRADIENT DECOMPOSITION ANALYSIS}
Based on the decomposition of $\nabla_{\bmu_i}L_j^k$ in main part, shown as follows:
\begin{equation}
    \frac{\partial L_j^k}{\partial \bmu_{i,x}^k} 
    = \frac{\partial L_j^k}{\partial c(\bp_j)}\times \frac{\partial c(\bp_j)}{\partial \alpha_i(\bp_j)}\times \frac{\partial \alpha_i(\bp_j)}{\partial \bmu_{i,x}^k}
    \label{eq:sup_gradeint1}
\end{equation}

\textit{\textbf{Gradient magnitude dilution.}} The first term $\partial L_j^k/\partial c(\bp_j)$ in above equation is only related to the loss function employed for training and irrelated with Gaussian primitives' parameters. Therefore, we can ignore it in following analysis. For clarity, we only focus on $x$-axis component and a single channel of rendering color in following analysis and similar conclusion can be drawn on $y$-axis component and remain two channels of rendering color. We can decompose the last two term of~\Cref{eq:sup_gradeint1} as follows (same as main part):
\begin{equation}
    \begin{aligned}
    & \frac{\partial c}{\partial \alpha _i} = c_i\prod_{j=1}^{i-1}(1-\alpha_j) + \sum_{l=i+1}^{N} (-c_l\alpha_l\prod_{j=1, j\neq i}^{l-1}(1-\alpha_j)) \\
    & \frac{\partial \alpha _i}{\partial \bmu_{i,x}^k} = o_i\times G_i^{2D}\times g'_{i,x},  \ \ \frac{\partial G_i^{2D}}{\partial \bmu_{i,x}^k} = G_i^{2D}\times g'_{i,x}
    \end{aligned}
    \label{eq:gradient2}
\end{equation}
Combined with above two equations, we can derive the $\frac{\partial c(\bp_j)}{\partial \alpha_i(\bp_j)}\times \frac{\partial \alpha_i(\bp_j)}{\partial \bmu_{i,x}^k}$ as follows:
\begin{equation}
    \begin{aligned}
    & \frac{\partial c}{\partial \alpha _i} \times \frac{\partial \alpha _i}{\partial \bmu_{i,x}^k} \\
    & =g'_x \left\{
        c_i o_i G_i^{2d}\prod_{j=1}^{i-1}(1-\alpha_j) \right. \\
    & \ \ \ \ \ \ \  \ \ \ \ \ \ \  \left. + o_i G_i^{2d} \sum_{l=i+1}^{N} \left[(-1) c_l\alpha_l\prod_{j=1, j\neq i}^{l-1}(1-\alpha_j)\right]
        \right\} \\ 
    & =g'_x \left\{
        c_i \alpha_i\prod_{j=1}^{i-1}(1-\alpha_j) - \alpha_i\sum_{l=i+1}^{N} \left[c_l\alpha_l\prod_{j=1, j\neq i}^{l-1}(1-\alpha_j)\right]
        \right\} \\
    & =g'_x \left\{
        c_i\omega_i - \alpha_i\sum_{l=i+1}^{N} \left[ c_l\alpha_l\prod_{j=1, j\neq i}^{l-1}(1-\alpha_j)\right]
        \right\} \\
    & =g'_x \left\{
        c_i\omega_i - \alpha_i\sum_{l=i+1}^{N} \left[ c_l\alpha_l\prod_{j=1}^{i-1}(1-\alpha_j)\prod_{j=i+1}^{l-1}(1-\alpha_j)\right]
        \right\} \\
    & =g'_x \left\{
        c_i\omega_i - \alpha_i\prod_{j=1}^{i-1}(1-\alpha_j)\sum_{l=i+1}^{N} \left[ c_l\alpha_l\prod_{j=i+1}^{l-1}(1-\alpha_j)\right]
        \right\} \\
    & =g'_x \left\{
        c_i\omega_i - \omega_i\sum_{l=i+1}^{N} \left[ c_l\alpha_l\prod_{j=i+1}^{l-1}(1-\alpha_j)\right]
        \right\}\\
    & =\omega_i \cdot \left\{
        c_i - \sum_{l=i+1}^{N} \left[\cdot c_l\alpha_l\prod_{j=i+1}^{l-1}(1-\alpha_j)\right]
        \right\} \cdot g'_x
    \end{aligned}
    \label{eq:importance_reasoning}
\end{equation}

Based on the above derivation, we can verify that $\partial L_j^k/\partial \bmu_{i,x}^k$ is positively correlated with the rendering importance score $\omega_i^k$. Therefore, overlooking $\omega_i^k$ in $\nabla \bmu_iL$ potentially causes gradient dilution. For a Gaussian primitive $G_i$ with high weight $\omega_i^k$ from the $k$-th viewpoint, $G_i$ likely appears earlier in the corresponding depth order. Consequently, $G_i$ may be occluded by other Gaussian primitives $G_j$ from different viewpoints, where $G_j$ plays a more significant role (located earlier in depth order), resulting in $G_i$ having smaller rendering weights in these viewpoints. Due to these low rendering weights leading to low gradient magnitudes, the overall average gradient magnitude of $G_i$ may fall below the threshold $\tau_{pos}$, as its high gradients from important viewpoints are diluted by numerous low gradients from less influential viewpoints. Under these conditions, $G_i$ will not be densified despite being located within complex regions that require more primitives to represent fine details. Our experiments confirm that this factor significantly impairs current 3D-GS optimization and contributes to blurring artifacts.

\noindent\textit{\textbf{Importance-aware densification.}} To overcome the gradient magnitude dilution, we propose the importance-aware densification criterion, which takes the rendering importance into account when normalizing the gradients from multiple viewpoints after several optimization iterations. For each pixel $\bp_j^k$ the $G_i$ is splatted onto from $k$ viewpoint, we need to multiply corresponding rendering weight $\omega_{i,j}^k$ with it as follows:
\begin{equation}
    \nabla_{\bmu_i}L^k_{j} = \omega_{i,j}^k \left(\frac{\partial L_j^k}{\partial \bmu_{i,x}^k}, \frac{\partial L_j^k}{\partial \bmu_{i,y}^k}\right),
\end{equation}
If $G_i$ is totally splatted on to $m_i^k$ pixels $\{\bp_j^k, j\in[1,\dots,m_i^k]\}$ in the $k$-th viewpoint, we need to aggregate all importance-aware $\nabla_{\bmu_i}L^k_j$ as the NDC gradient of $G_i$ (consistent with 3D-GS rendering engine) as follows:
\begin{equation}
    \nabla_{\bmu_i}L^k = \left(\sum_j^{m_i^k}\omega_{i,j}^k\cdot\frac{\partial L_j^k}{\partial \bmu_{i,x}^k}, \sum_j^{m_i^k}\omega_{i,j}^k\cdot\frac{\partial L_j^k}{\partial \bmu_{i,y}^k}\right)
\end{equation}
which leads to additional memory consumption and computational overhead in parallel CUDA kernel. Therefore, we simplify the above formulation as follows:
\begin{equation}
    \nabla_{\bmu_i}L^k = \omega_{i}^k\cdot\left(\sum_j^{m_i^k}\frac{\partial L_j^k}{\partial \bmu_{i,x}^k}, \sum_j^{m_i^k}\frac{\partial L_j^k}{\partial \bmu_{i,y}^k}\right), \ \ \omega_{i}^k = \frac{\sum_j^{m_i^k}\omega_{i,j}^k}{m_i^k}
\end{equation}
Notably, the $\omega_{i}^k$ can be easily accumulated and calculated with only one additional \textit{float} memory consumption and one additional \textit{int} memory to record $m_i^k$. Meanwhile, the normalization of $\omega_{i}^k$ over projection area $m_i^k$ also eliminates the bias caused by large-scale Gaussians being split aggressively, thus preventing the geometric artifacts illustrated in main part. Based on the new importance-aware $\nabla_{\bmu_i}L^k$, we can obtain importance-aware $\|\nabla_{\bmu_i}L^k\|$ as:
\begin{equation}
\begin{aligned}
    \|\nabla_{\bmu_i}L^k\| & = \omega_{i}^k\cdot\sqrt{\left(\frac{\partial L^k}{\partial \bmu_{i,x}^k}\right)^2 + \left(\frac{\partial L^k}{\partial \bmu_{i,y}^k}\right)^2 }\\
    \frac{\partial L^k}{\partial \bmu_{i,x}^k} & = \sum_j^{m_i^k}\frac{\partial L_j^k}{\partial \bmu_{i,x}^k} \\
    \frac{\partial L^k}{\partial \bmu_{i,y}^k} & = \sum_j^{m_i^k}\frac{\partial L_j^k}{\partial \bmu_{i,y}^k}
\end{aligned}
\end{equation}
and then we redefine the importance-aware densification criterion $\nabla_{\bmu_i}L$ as:
\begin{equation}
    \nabla{\bmu_i}L = \frac{\sum_{k=1}^{M}\|\nabla_{\bmu_i}L^k\|}{\sum_{k=1}^{M}\omega_i^k}
    =\frac{\sum_{k=1}^{M} \omega_i^k\sqrt{\left(\frac{\partial L^k}{\partial\bmu_{i,x}^k}\right)^2 + \left(\frac{\partial L^k}{\partial\bmu_{i,y}^k}\right)^2}}{\sum_{k=1}^{M}\omega_i^k}.
\end{equation}

\section{ADDITIONAL RESULTS}

\begin{figure*}[t]
    \centering
    \includegraphics[width=0.95\linewidth]{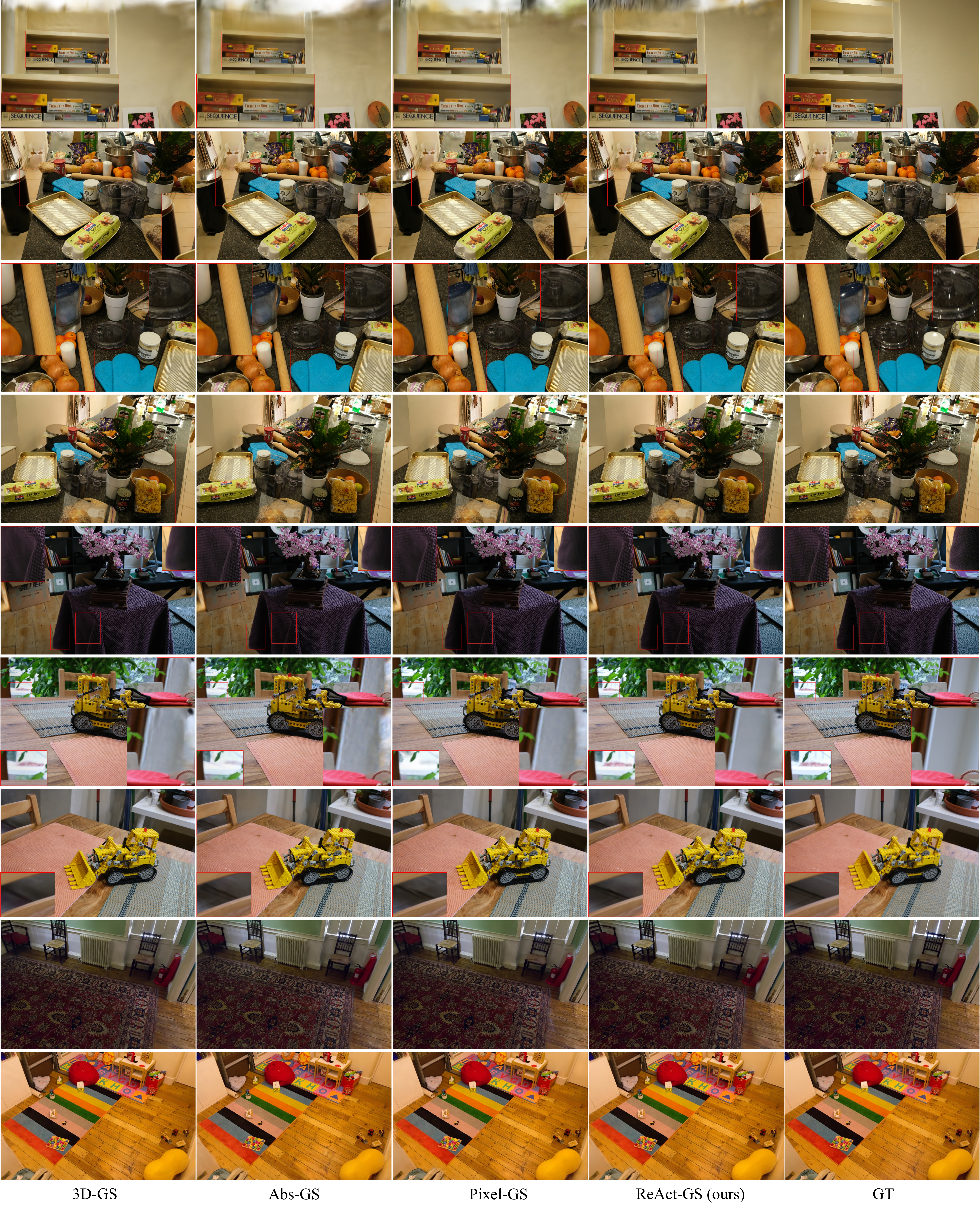}
    \caption{Qualitative comparisons of different methods on indoor scenes.}
    \label{fig:sm_indoor}
\end{figure*}

\begin{figure*}[t]
    \centering
    \includegraphics[width=\linewidth]{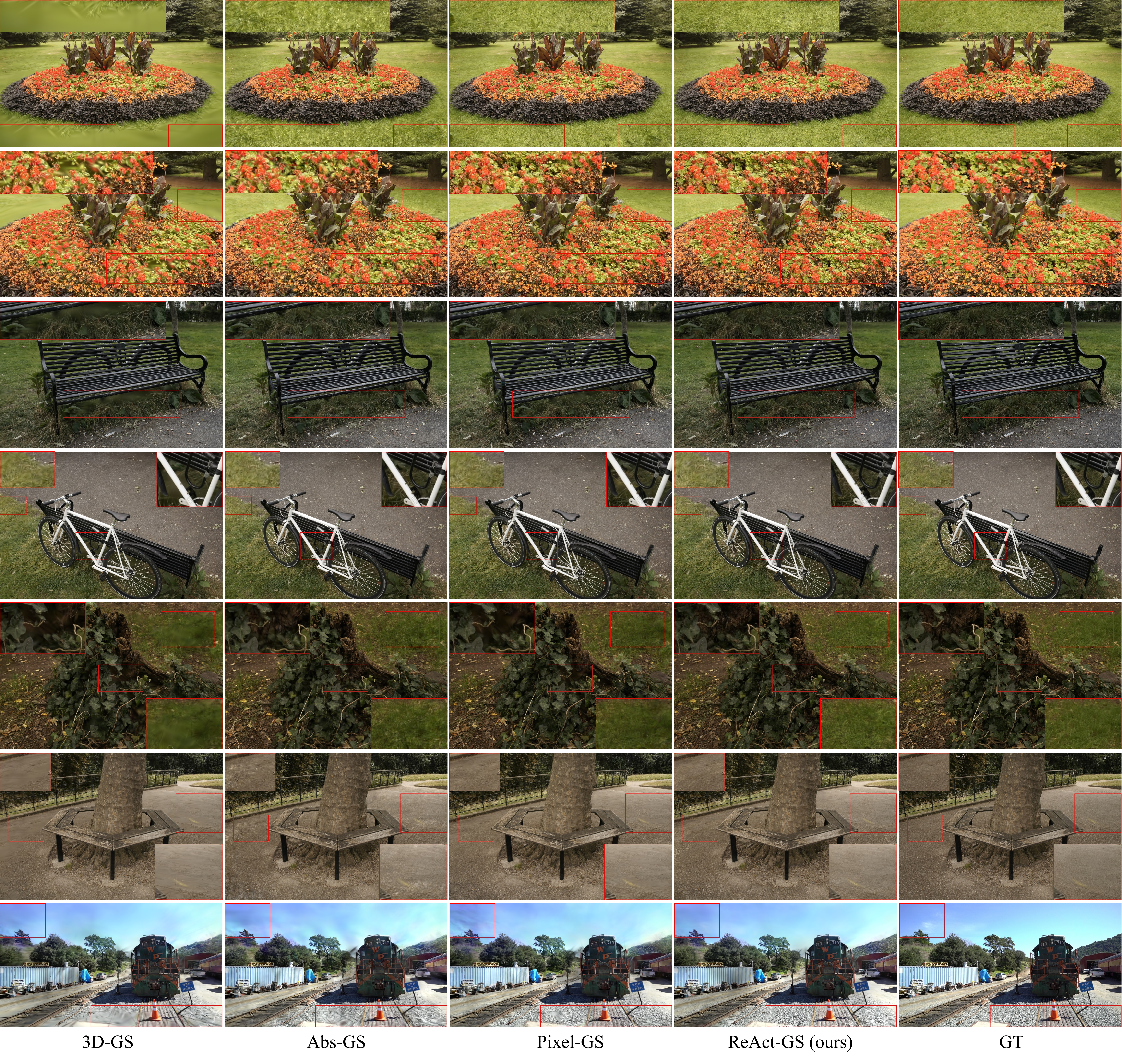}
    \caption{Qualitative comparisons of different methods on outdoor scenes.}
    \label{fig:sm_outdoor}
\end{figure*}

\begin{figure*}[t]
    \centering
    \includegraphics[width=\linewidth]{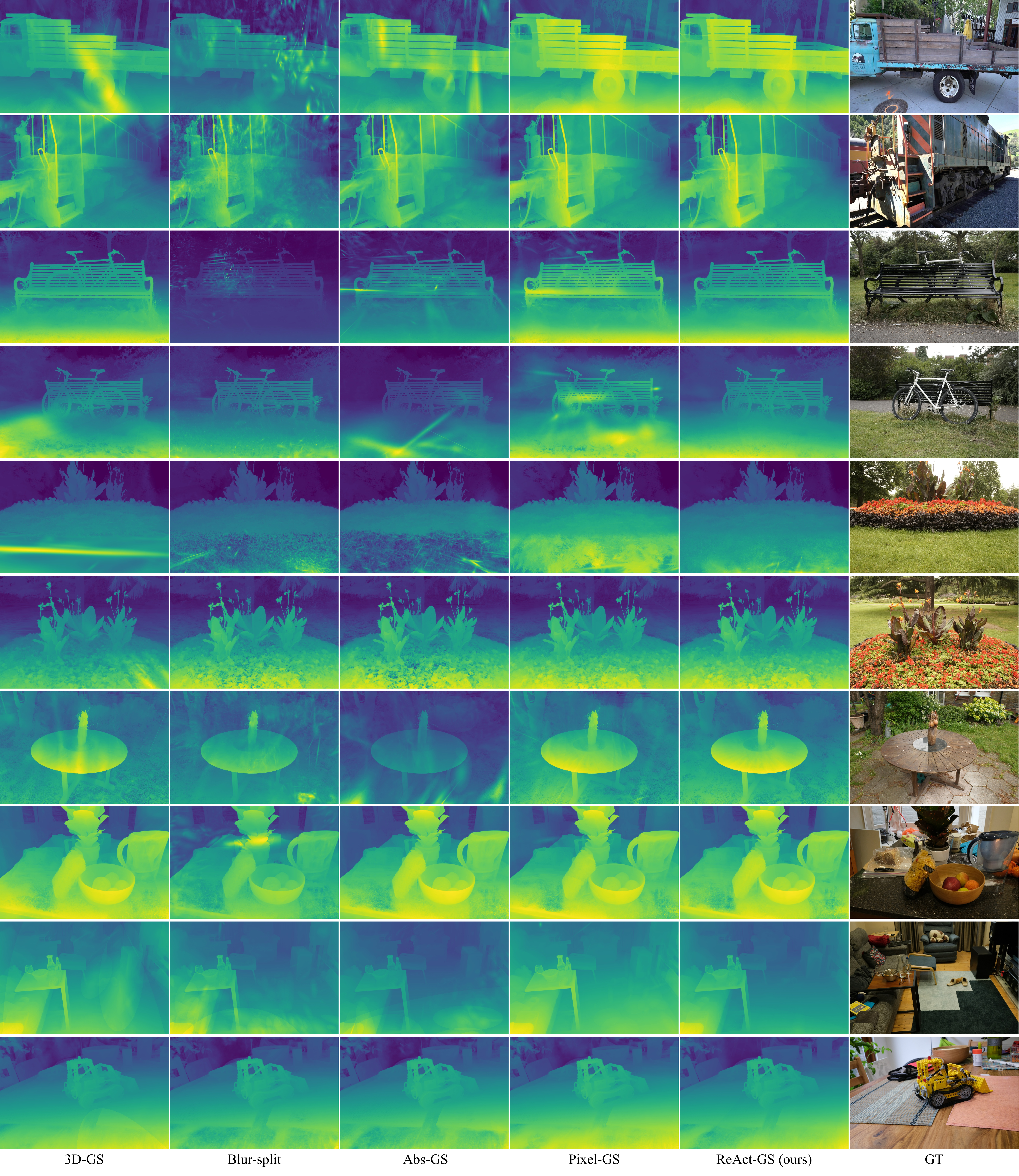}
    \caption{Depth visualization of different methods on various scenes.}
    \label{fig:sm_depth}
\end{figure*}

\noindent\textit{\textbf{Additional qualitative experiments results.}}
~\Cref{fig:sm_indoor} and ~\Cref{fig:sm_outdoor} provide more visualization results of baselines and our methods. ~\Cref{fig:sm_depth} provide more visualizations on depth maps of baselines and our methods

\end{document}